%% file: ms.tex
\documentclass[final]{cvpr}

\usepackage{times}
\usepackage{epsfig}
\usepackage{graphicx}
\usepackage{amsmath}
\usepackage{amssymb}
\usepackage{multirow}
\usepackage{siunitx}
\usepackage{caption}
\usepackage{subcaption}

\usepackage{color}
\definecolor{patrick_color}{rgb}{.6,.4,.05}
\definecolor{chengcheng_color}{rgb}{.5,.7,.1}
\definecolor{chris_color}{rgb}{0,0.35,0}
\definecolor{mv_color}{rgb}{0,0,1}
\definecolor{charlie_color}{rgb}{0,0,0.8}
\definecolor{samarth_color}{rgb}{0.75,0.25,0.0}

\newcommand{\twopartdef}[4]
{
	\left\{
		\begin{array}{ll}
			#1 & #2 \\
			#3 & #4
		\end{array}
	\right.
}

\usepackage{hyperref}
\hypersetup{pagebackref=true,breaklinks=true,colorlinks,bookmarks=false}

\def\cvprPaperID{3479} 
\def\confYear{CVPR 2021}

\begin{document}
\newcommand{\algoName}{ContactOpt\xspace}
\newcommand{\networkName}{DeepContact\xspace}
\newcommand{\meshContactName}{DiffContact\xspace}

\title{ContactOpt: Optimizing Contact to Improve Grasps}

\author{Patrick Grady\textsuperscript{1}, Chengcheng Tang\textsuperscript{2}, Christopher D. Twigg\textsuperscript{2}, Minh Vo\textsuperscript{2}, \\Samarth Brahmbhatt\textsuperscript{3}, Charles C. Kemp\textsuperscript{1}\\
\\
\textsuperscript{1}Georgia Institute of Technology, \textsuperscript{2}Facebook Reality Labs Research, \textsuperscript{3}Intel Labs\\



}

\twocolumn[{%
\renewcommand\twocolumn[1][]{#1}%
\maketitle

\begin{center}
\centering
\vspace{-0.7cm}
\includegraphics[width=15.5cm]{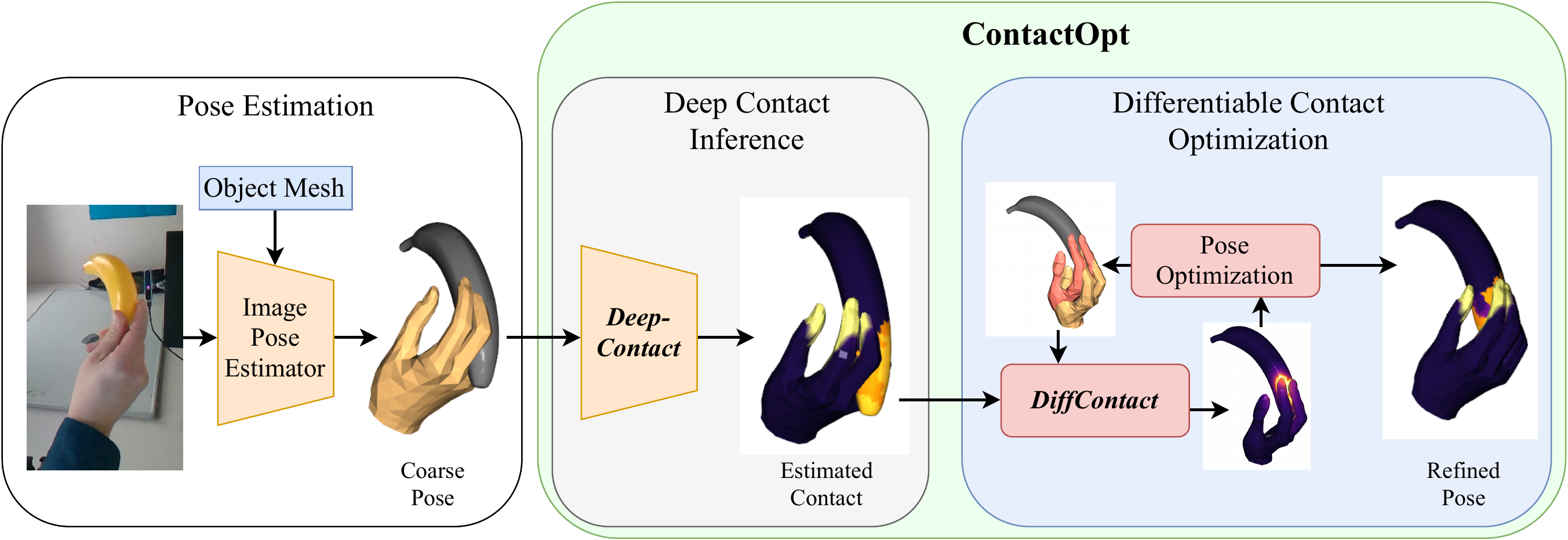}
\end{center}
\vspace{-0.3cm}
Figure 1: \algoName pipeline. Left: A pose estimator generates a hand pose. Middle: \networkName estimates where contact should occur (target contact). Right: The hand pose is optimized to achieve target contact via a contact model (\meshContactName).
\vspace{3mm}
}]

\setcounter{figure}{1}  

\begin{abstract}
\vspace{-3mm}
Physical contact between hands and objects plays a critical role in human grasps. We show that optimizing the pose of a hand to achieve expected contact with an object can improve hand poses inferred via image-based methods. Given a hand mesh and an object mesh, a deep model trained on ground truth contact data infers desirable contact across the surfaces of the meshes. Then, ContactOpt efficiently optimizes the pose of the hand to achieve desirable contact using a differentiable contact model. Notably, our contact model encourages mesh interpenetration to approximate deformable soft tissue in the hand. In our evaluations, our methods result in grasps that better match ground truth contact, have lower kinematic error, and are significantly preferred by human participants. Code and models are available online\footnote{\url{https://github.com/facebookresearch/contactopt}}.

\end{abstract}



\input{1-introduction}

\input{2-related-works}

\input{3-approach}

\input{4-results}

\input{5-conclusion}

{\small
\bibliographystyle{ieee_fullname}
\bibliography{egbib}
}

\end{document}

%% file: 1-introduction.tex

\vspace{-0.5cm}

\section{Introduction} \label{sec:intro}





The availability of data, hand and body models, and learning algorithms has fueled a growing interest in capturing, understanding, and simulating hand-object interactions \cite{Brahmbhatt_2020_ECCV, Glauser:Stretch-Glove:2019, hampali2020honnotate, sundaram2019learning, GRAB:2020, Freihand2019}. Recent algorithms can predict hand and object pose increasingly accurately from an image. However, inferred poses continue to exhibit sufficient error to cause unrealistic hand-object contact, making downstream tasks in simulation, virtual reality, and other applications challenging.

A key issue is that physical contact is sensitive to small changes in pose. For example, less than a millimeter change in the pose of a fingertip normal to the surface of an object can make the difference between the object being held or dropped on the floor. In addition to physical implausibility, lack of contact and other small-scale phenomena can reduce the perceptual realism of rendered poses. 

In this paper we present ContactOpt, an algorithm that improves the quality of hand-object contact by refining hand pose. When given a hand mesh and an object mesh, ContactOpt infers where contact is likely to occur and then optimizes the hand pose to achieve this contact. 


As shown in Figure 1, ContactOpt consists of two main components, \networkName and \meshContactName. \networkName is a network that takes the hand and object meshes as input and estimates regions of likely contact. \meshContactName is a differentiable function that takes the hand and object meshes as input and outputs contact based on current geometry. ContactOpt uses gradient-based optimization to find pose, translation, and rotation parameters for the MANO hand model \cite{romero2017embodied} that improve the match between current contact from \meshContactName and target contact from \networkName.


Notably, ContactOpt takes into account soft tissue deformation in the hand. The inner surface of a human hand undergoes significant deformation when making contact with objects. For example, the finger pad can deform 2-3 mm, and the palm can deform 5 mm under normal grasping forces ~\cite{perez2013stiffness}. \meshContactName permits up to 2 mm of interpenetration between the hand and object meshes without penalty. In addition, ContactOpt's gradient-based optimization uses a loss function that only penalizes penetration greater than this threshold. This allows for contact to occur across wide areas of the hand, rather than only at single points.



We conducted two types of evaluations to assess ContactOpt's performance. For the first type of evaluation, we evaluated ContactOpt's ability to refine hand pose estimates with \emph{small} inaccuracies in dataset annotations. This presents methodological challenges due to limits in the precision of dataset ground truth annotations. To overcome this, we used the ContactPose dataset, which has both pose estimates \textit{and} measured contact data obtained via thermal imagery. We had ContactOpt refine these hand pose estimates with respect to ground truth contact. The refined hand poses better matched ground truth contact and were preferred by human participants, demonstrating that ContactOpt can improve state-of-the-art pose estimates from existing datasets.

\begin{figure}[t]
\begin{center}
  \includegraphics[width=0.8\linewidth]{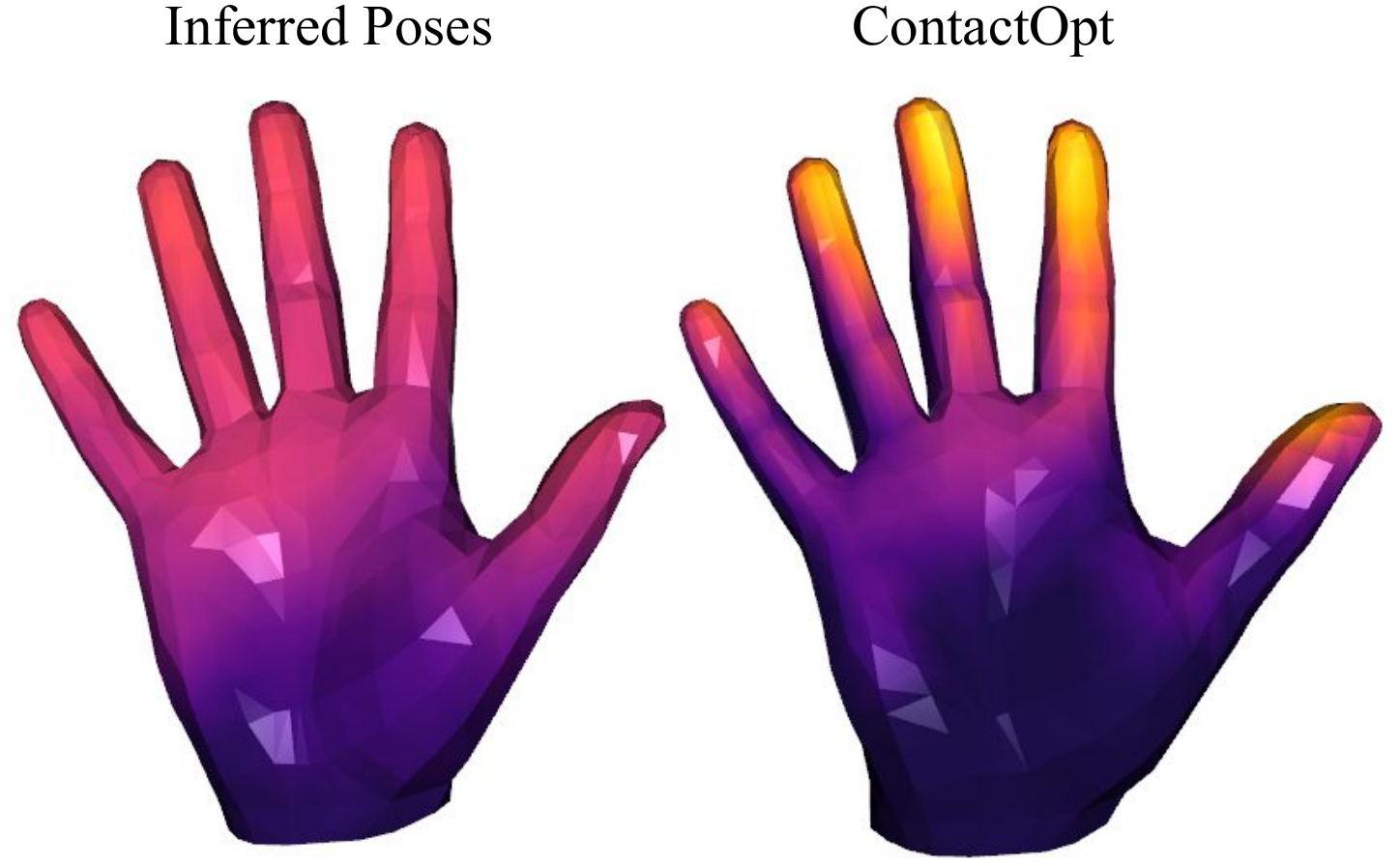}
\end{center}
\vspace{-12pt}
  \caption{Frequency of hand contact calculated with poses inferred with an image-based pose estimator \cite{hasson2020leveraging} (left) and after refinement with ContactOpt (right).  Note the increase in contact on the finger pads and across the index finger.}
  \label{fig:handContactFreq}
\end{figure}

For the second type of evaluation, we evaluated ContactOpt's ability to refine hand pose estimates with \emph{large} inaccuracies.
We used ContactOpt to refine hand pose estimates from an existing RGB hand pose estimation network (Hasson et al. \cite{hasson2020leveraging}) applied to the HO-3D dataset \cite{hampali2020honnotate}. ContactOpt's refined hand poses had lower kinematic error, were preferred by human participants, and matched more closely to previously observed hand contact patterns (Figure \ref{fig:handContactFreq}). ContactOpt also outperformed RefineNet \cite{GRAB:2020} (an end-to-end grasp refinement network) with respect to both measures. This demonstrates ContactOpt's value as a post-processing stage for existing hand-object pose estimation algorithms for which it has not been specifically trained. Since ContactOpt operates on hand and object meshes, it has the potential to improve the output of recent image-based estimation methods 
while avoiding some types of generalization issues associated with operating on images. 


\vspace{3mm}
In summary, our contributions follow: 
\begin{itemize}
    \setlength\itemsep{0.1mm}
    \item We show that methods that explicitly consider hand-object contact can improve hand pose estimates at both coarse ($\approx$cm) and fine ($\approx$mm) spatial scales, resulting in improved visual realism and lower kinematic error. 
    
    
    
    
    \item We present \networkName, a deep network that estimates where contact is likely to occur across the surfaces of inaccurately aligned hand and object meshes.
    
    \item We present \meshContactName, a differentiable contact model that estimates where contact is occurring between hand and object meshes.
    
    \item We present ContactOpt, an algorithm that performs gradient-based optimization to improve hand-object contact by refining hand pose.

\end{itemize}

%% file: 2-related-works.tex
\section{Related Work} \label{sec:related_work}
In this work, we use likely contact and a contact model to improve the pose of a hand grasping an object.
Applications in computer vision, animation, and robotics have driven interest in hand-object interaction tracking from different angles, e.g., recovering poses from input images or generating grasps based on object pose and geometry.
Information about contact is playing an increasingly important role for hand-object interaction tracking, grasp generation and multiple other related applications.

\paragraph{Datasets of hand-object contact.}

Recently, there has been a focus on collecting datasets that include interactions between hands and objects.  FreiHand~\cite{Freihand2019} 
uses multiple cameras to extract high-quality annotations using the MANO model, but does not include the object pose.  
HO-3D~\cite{hampali2020honnotate} optimizes simultaneously for both hand and object poses from RGB-D sensors.  FHAB~\cite{garcia2018first} leverages a unique magnetic tracking system to infer the pose of a hand and object even under occlusion.  GRAB~\cite{GRAB:2020} uses professional optical motion capture to collect a dataset of people grasping and manipulating objects. 
The work additionally infers contact from the proximity of hand and object. However, these estimates may be noisy due to the very high pose accuracy necessary to infer accurate contact.


Datasets for contact directly measured on objects~\cite{Brahmbhatt_2019_CVPR, lau2016tactile} and hands~\cite{sundaram2019learning} are complementary to datasets on hand-object poses.
The ContactPose dataset \cite{Brahmbhatt_2020_ECCV} is unique in capturing both ground truth thermal contact maps, as well as hand and object pose.  The participants held a static grasp for each of 25 objects while being captured using multiple RGB-D cameras. The object was tracked using motion capture, and the hand pose was estimated by aggregating predictions across time from an RGB hand pose estimator.  A thermal camera measured the body heat transferred from the participant's hand to the object, providing ground truth contact.  The dataset shows that contact occurs across large sections of the hand, as opposed to only at the fingertips. A limitation of the method is that the 3D hand pose accuracy is bounded by the accuracy of the hand pose estimation, so there may be discrepancies between the contact map and the MANO hand mesh.

\paragraph{Image-based hand-object pose estimation.}

There is an extensive body of work on estimating the pose of the hand using a variety of input modalities, including: gloves with markers or sensors~\cite{Glauser:Stretch-Glove:2019, han2018online, wang2009real}, depth/RGB-D input~\cite{bhatnagar2020combining, EgoDexterDataset, DexterWithObjectDataset, Dexter1Dataset, MSRADataset, tagliasacchi2015robust, ICVLDataset, tkach2016sphere, NYUDataset, bigHand}, and RGB or monochrome images~\cite{affinityFields, doosti2020hope, ganeratedHandsDataset, spurr2020weakly, bigHand, Freihand2019},
with an increasing focus on hand-object interaction~\cite{ doosti2020hope, hamer2009tracking, hampali2020honnotate, hasson2020leveraging, hasson2019learning,karunratanakul2020grasping, oikonomidis2011full,panteleris20153d,romero2010hands,DexterWithObjectDataset, tekin2019ho}.
Researchers have long realized that inferring and enforcing contact is important for hand-object interaction tracking~\cite{romero2010hands, tzionas2016capturing}, and it remains a challenging task, particularly in the absence of depth data.
For RGB-D hand tracking, hand-object contact modeled as finger-tip to object distance was part of the energy function during optimization with Gaussian Mixture Models in~\cite{DexterWithObjectDataset}.
For image-based prediction, skeletal hand poses~\cite{doosti2020hope, tekin2019ho} or MANO~\cite{romero2017embodied} hand model parameters~\cite{hasson2020leveraging, hasson2019learning} are predicted jointly with object geometry or pose in an end-to-end manner.
Despite sharing a joint latent space, since the output representations for the hand and object are decoupled, there can be \textit{relative} errors in the poses, leading to unrealistic grasps.
Even though contact can be encouraged at training time, these networks have no method of \textbf{enforcing} alignment at test time. 
Our work complements these existing methods by leveraging the strength of their joint hand-object pose prediction, but uses explicit contact inference and enforcement to achieve higher quality grasps.

\paragraph{Grasp synthesis.}

Robotic grasp generation shares many similarities to pose refinement. Generally, the robot attempts to find a stable grasp with high robustness to perturbations. Various input modalities have been explored for learned grasp detectors, including depth \cite{mahler2017dexnet, morrison2018closing} and RGB \cite{chu2018real, le2010learning, pinto2016supersizing, saxena2008robotic}. Some methods use physics simulation \cite{depierre2018jacquard, graspit} or analytical heuristics \cite{rosales2011global} to find stable grasps. The majority of robotic grasping work focuses on simple grippers with sparse contact points, however some research has investigated manipulation with anthropomorphic hands \cite{openai2020rl}.

Similarly, generating plausible grasps for a human hand has also been explored. 
In GanHand~\cite{corona2020ganhand}, a dataset of affordances and grasps was proposed to generate plausible human grasps based on input images. The works that are most similar to ours are ContactGrasp~\cite{brahmbhatt2019contactgrasp} and GRAB~\cite{GRAB:2020}. In ContactGrasp~\cite{brahmbhatt2019contactgrasp}, dense ground truth contact maps from ContactDB are used to generate plausible grasps for a given object geometry. However, this requires pre-recorded contact maps, and because the ContactDB dataset lacks ground truth hand poses, they cannot compare against ground truth or condition on images as we do.
In GRAB~\cite{GRAB:2020} the authors leverage their collected data to generate compelling grasps for a variety of objects.  They propose RefineNet, which improves the quality of a grasp given an initial pose. This has similarities to our approach, but it performs end-to-end pose updates rather than optimization, and considers fixed contact patterns as opposed to contact estimated separately for each grasp. The method does not explicitly consider object geometry, and because it is fully learned, may have less ability to generalize.  We show comparisons against this approach when applied to image-based inference tasks in \secref{sec:experiments}.

\paragraph{Contact in human pose.}

Aside from hand-object interaction, contact is informative for full human body poses given human-environment interaction~\cite{clever2020bodies, narasimhaswamy2020detecting}.
Inferred contact constraints are used in~\cite{rempe2020contact} to improve body pose estimation from videos to mitigate artifacts such as feet sliding.
Coarse contact points are used in generating human poses interacting with scenes~\cite{holden2020learned, starke2020local, zhang2020generating}.
Our work leveraging fine-grain contact information to improve hand pose in hand-object interaction tracking is related to and likely applicable to context-aware full-body pose estimation and generation.

%% file: 3-approach.tex
\section{Methods}

\newcommand{\ctop}[0]{\ensuremath{c_{\text{top}}}}
\newcommand{\cbot}[0]{\ensuremath{c_{\text{bot}}}}
\newcommand{\crad}[0]{\ensuremath{c_{\text{rad}}}}

\begin{figure*}
\begin{center}
   \includegraphics[width=1.0\linewidth]{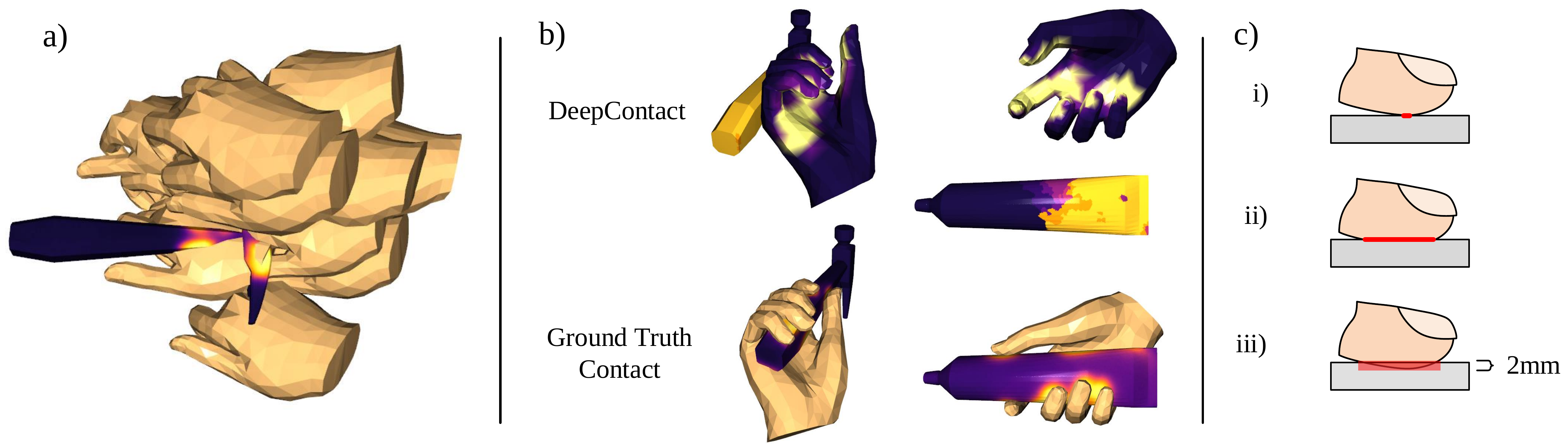}
\end{center}
\vspace{-8pt}
\caption{a) Example of multiple hand poses from Perturbed ContactPose, all generated from a single dataset sample. b) Top: \networkName predicts contact maps for the hand and object as if they were aligned. Bottom: Ground truth poses and thermal contact. c) When a human finger contacts an object, point contacts are rare (i). More commonly, the soft tissue in the finger conforms to the surface (ii) resulting in a large area of contact. While the MANO mesh does not locally deform to match the surface, we can encourage the optimizer to create matching area-based contact by marking vertices as being in contact even when they are 2 mm inside the surface (iii).}
\label{fig:page4fig}
\end{figure*}

We represent the grasp with an object mesh $\mathbf{O}$ and a MANO \cite{romero2017embodied} hand mesh $\mathbf{H}$. $\mathbf{H}$ is described by parameters $\mathbf{P} = \left(\theta, \beta, {t}^H, {R}^H\right)$, consisting of pose, shape, translation, and rotation w.r.t. the object respectively. Pose $\theta$ is represented as a 15-dimensional PCA manifold, which lowers the high-dimensional joint angle representation to a compressed space of typical hand poses.

Given a noisy estimate of $\mathbf{P}$ (which typically comes from an image-based algorithm), we seek a better grasp by exploiting the hand-object contact information. Figure 1 shows an overview of our approach. In the following sections, we describe our learned contact map estimation module \networkName (Section \ref{sec:contact_inference}) and our differentiable contact model \meshContactName (Section \ref{sec:diff_contact_model}) that is iteratively updated according to the optimized hand pose to reproduce the estimated contact (Section \ref{sec:contact_optim}).

\subsection{\networkName: Learning to Estimate Contact} \label{sec:contact_inference}

Given an object mesh $\mathbf{O}$ and and hand mesh $\mathbf{H}$ with potentially inaccurate pose $\mathbf{P}$, \networkName{} learns to infer target contact on the hand $\hat{C}_H$ and object $\hat{C}_O$. 



We represent the meshes $\mathbf{H}$ and $\mathbf{O}$ as point clouds, and use PointNet++ \cite{qi2017pointnet++} to predict contact. The object pointcloud contains 2048 points randomly sampled from the object. The hand point cloud contains all 778 vertices of the MANO mesh. We employ the ``mesh" features, training loss, and discrete contact representation of Brahmbhatt et al.~\cite{Brahmbhatt_2020_ECCV}. The ``mesh" features capture distances from the hand to the object, as well as normal information. Additionally, we include a binary per-point feature indicating whether the point belongs to the hand or the object. The network predicts contact as a classification task, where the range $[0, 1]$ is split into 10 bins. We train DeepContact with the standard binary cross-entropy loss.


Similarly to GrabNet \cite{GRAB:2020}, we train this module on a dataset of randomly perturbed hand poses from the ContactPose dataset, which we call \textbf{Perturbed ContactPose}. The hand mesh is modified by adding noise to the parameters $\Delta\theta \sim \mathcal{N}(0, 0.5)$, $\Delta {t}^H \sim \mathcal{N}(0, 5)$ cm, and $\Delta {R}^H \sim N(\ang{0}, \ang{15})$. Object contact is supervised with ground-truth thermal contact from ContactPose. To generate the target hand contact map, we run \meshContactName (section \ref{sec:diff_contact_model}). By applying multiple perturbations to each grasp, a training/testing split of 22K / 1.4K grasps is generated.

Figure~\ref{fig:page4fig}a shows example perturbations, and Figure \ref{fig:page4fig}b shows an example contact prediction. Hand and object poses that are farther from a particular grasp tend to result in larger and more diffuse areas of predicted contact.

\begin{figure*}
\begin{center}
   \includegraphics[width=1.0\linewidth]{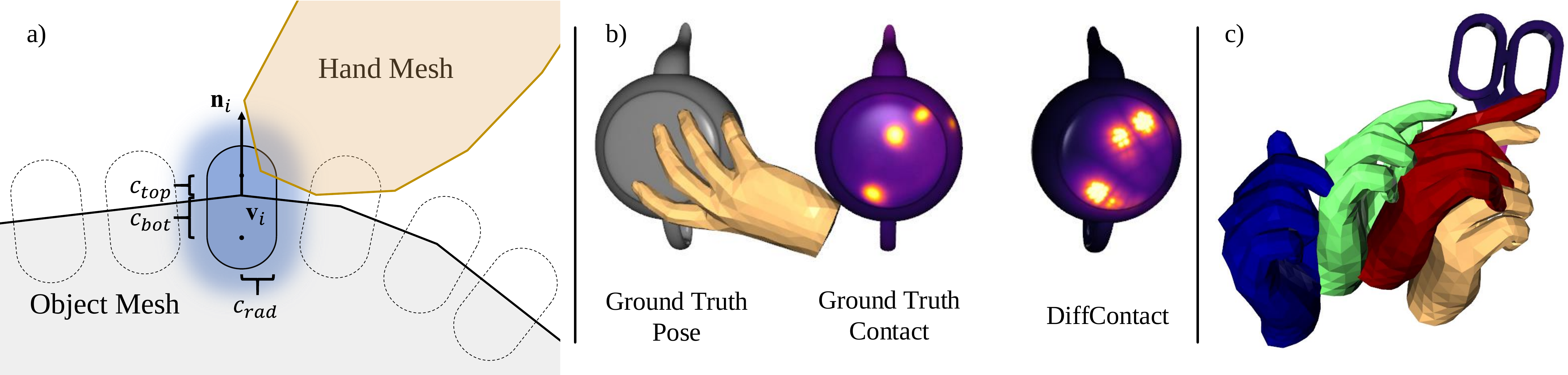}
\end{center}
\vspace{-8pt}
\caption{a) Virtual capsules are placed on each vertex of the object, aligned with the vertex normal. If any hand vertices are inside the capsule, the object point is marked as being in full contact. b) Left: Hand and object from ContactPose dataset. Center: Ground truth thermal contact. Right: Contact estimated from \meshContactName{}. c) Optimization of hand pose to match target contact. From left to right: hand pose at selected iterations during optimization.}
\label{fig:page5fig}
\end{figure*}

\subsection{\meshContactName{}: Differentiable Contact Model} \label{sec:diff_contact_model}

\meshContactName{} estimates the contact maps, $C_O(\mathbf{P})$ and $C_H(\mathbf{P})$, based on the current meshes $\mathbf{O}, \mathbf{H}(\mathbf{P})$. This is done in a differentiable way, allowing optimization of the hand parameters $\mathbf{P}$.

We propose a contact model using \textit{virtual capsules}, as shown in Figure~\ref{fig:page5fig}a. Our virtual capsules have useful attraction extended beyond the surface (which a binary proximity would not) and approximate soft hand tissue deformation. 

More concretely, we place a virtual capsule at every object vertex $\mathbf{v}^O_i$ and orient it along the object surface normal $\mathbf{n}^O_i$. This capsule has a principal line segment defined by $\mathbf{v}^O_i + \alpha \mathbf{n}^O_i$, $\alpha \in [-\cbot, \ctop]$.
Let $\phi(\mathbf{x})$ be the Euclidean distance from a 3D point $\mathbf{x}$ to this line segment.
The contact is defined to be uniformly 1 for points such that $\phi(\mathbf{x}) < \crad{}$ and falls off proportionally with distance outside $\crad{}$ as $\frac{ \crad}{\phi(\mathbf{x})}$.

Let $\mathbf{v}_j^H(\mathbf{P})$ be the hand vertex at pose $\mathbf{P}$ with the smallest distance $\phi$ to the object vertex $\mathbf{v}_i^O$. The contact value at the object vertex $\mathbf{v}_i^O$ is expressed as:

\begin{equation}
C_O\left(\mathbf{v}^O_i; \mathbf{P}\right) = \min \left(\frac{ \crad}{\phi(\mathbf{v}^H_j(\mathbf{P}))}, 1\right).
\end{equation}

The same procedure can be used to calculate the contact map on the hand surface. We choose an asymmetric $\cbot > \ctop$ such that the region considered ``in contact'' extends farther inside the mesh than outside, which approximates soft hand tissue deformation as shown in Figure~\ref{fig:page4fig}c. In our experiments, $\ctop = 0.5$ mm, $\cbot = 1$ mm, and $\crad = 1$ mm.   As the total capsule depth inside the object is $\cbot{} + \crad{} = 2$ mm, this conservatively matches the $2-3$ mm finger pad deformation found in the biomechanics literature~\cite{cabibihan2014illusory,contactMechanicsFingerPad}.

Figure~\ref{fig:page5fig}b shows an example of object contact computed with this model. Because the generated contact has a gradual dropoff, this provides gradients for optimization. Additionally, the resulting contact maps have diffuse edges, which appear visually similar to thermal contact maps \cite{Brahmbhatt_2019_CVPR, Brahmbhatt_2020_ECCV}. The generated contact is an \textit{area} instead of a single point.

\subsection{Contact Optimization} \label{sec:contact_optim}
To align the meshes, the hand mesh parameters $\mathbf{P}$ are iteratively optimized (Figure \ref{fig:page5fig}c) to minimize the difference between the current contact maps $C_H(\mathbf{P})$, $C_O(\mathbf{P})$ computed using \meshContactName{}, and the target contact maps $\hat{C}_H$, $\hat{C}_O$ as predicted by DeepContact, or from ground truth thermal contact.

The contact loss for the object surface is:
\begin{equation}
E_O(\mathbf{P}) = 
    \twopartdef{\lambda |C_O(\mathbf{P}) - \hat{C}_O|}
                    {\mbox{if } C_O(\mathbf{P}) < \hat{C_O}}
               {|C_O(\mathbf{P}) - \hat{C}_O)|}{\text{otherwise}}
\end{equation}
Here we use $\lambda > 1$ to penalize ``missing'' contacts (where the target contact is higher than the value estimated by \meshContactName) more heavily than ``unexpected'' contacts. This is based on the empirical observation that it is visually worse for the hand to ``hover'' over the object than to be slightly interpenetrating.   We apply a corresponding loss $E_H(\mathbf{P})$ to penalize differences between the target hand contact map $\hat{C}_H$ and $C_H(\mathbf{P})$.  We use $\lambda=3$ in both cases.

We also include an explicit penetration term that penalizes penetrations beyond $c_{pen}$. This discourages heavy intersection where vertices on the back of the hand register as in contact. For each object vertex $\mathbf{v}^O_i$, object surface normal $\mathbf{n}^O_i$, and nearest hand vertex $\mathbf{v}^H_j(\mathbf{P})$, the penetration loss is defined as
\begin{equation}
E_{pen}(\mathbf{P}) = \sum_{i} \text{max}\left(0, (\mathbf{v}^O_i-\mathbf{v}^H_j(\mathbf{P})) \cdot \mathbf{n}^O_i - c_{pen}\right)
\end{equation}
where $c_{pen} = 2$ mm. The final loss is 
\begin{equation}
E(\mathbf{P}) = E_H(\mathbf{P}) + \lambda_O E_O(\mathbf{P}) + \lambda_{pen} E_{pen}(\mathbf{P}) 
\end{equation}


The loss is minimized by the ADAM optimizer \cite{adam} using gradients computed with PyTorch automatic differentiation~\cite{pytorch}. We use a learning rate of $0.01$ and optimize for 250 iterations. Optimizing a batch of 64 hand-object pairs takes 4 s (amortized runtime 62 ms). We scale the gradients for the different components of $\mathbf{P}$. See the supplementary material for more details.

\paragraph{Random restarts.} Since the contact optimization is local, a poor initialization (\eg initial hand position on the wrong side of an object) can result in the optimizer settling into a bad local minimum.  We avoid this by applying the pose optimization to several perturbations of the provided pose and select the result with the lowest loss.

%% file: 4-results.tex
\section{Evaluation} \label{sec:experiments}

We evaluate how well \algoName improves poses with small inaccuracies and with large inaccuracies using the ContactPose and HO-3D datasets. In each case, the refined hand mesh is evaluated using the following metrics.

\begin{itemize}
    \item \textbf{Intersection Volume} (cm$^3$): Intersection volume of $\mathbf{H}$ and $\mathbf{O}$, calculated from their mesh intersection. Standard deviation across the dataset is also shown.
    
    \item \textbf{Mean Per-Joint Position Error (MPJPE)} (mm): Average L2 per-joint kinematic error with respect to the ground truth hand~\cite{h36m_pami}.
    
    \item \textbf{Contact Coverage} (\%): Percentage of hand points between -2 mm and +2 mm of the object surface (i.e., approximately in contact with the object).
    
    \item \textbf{Contact Precision/Recall} (\%): Quantifies how well the contact from the refined hand mesh matches the thermal contact map. A binary object contact map is obtained by considering the object points within $\pm$2 mm of the hand surface to be in contact. Precision and recall are calculated by comparing this to the thermal contact map thresholded at 0.4, following \cite{Brahmbhatt_2020_ECCV}.
    
    \item \textbf{Perceptual Evaluation} (\%): Nine evaluators who were unfamiliar with the research were recruited to judge the relative quality of grasps in two-alternative forced choice tests (2AFC). Each participant was shown two hand-object pairs and asked to judge ``Which looks more like the way a person would grasp the object?". In pilot studies, we found that non-experts had difficulty comparing grasps with small differences, so pairs with less than a 5 mm MPJPE difference were removed. For each method, the evaluators judged 75 pairs of grasps with an equal number randomly selected for each object. The mean and 95\% confidence intervals are shown. More details of this evaluation can be found in the supplementary material.
    
    
    
\end{itemize}



\begin{figure}
\begin{center}
   \includegraphics[width=1.0\linewidth]{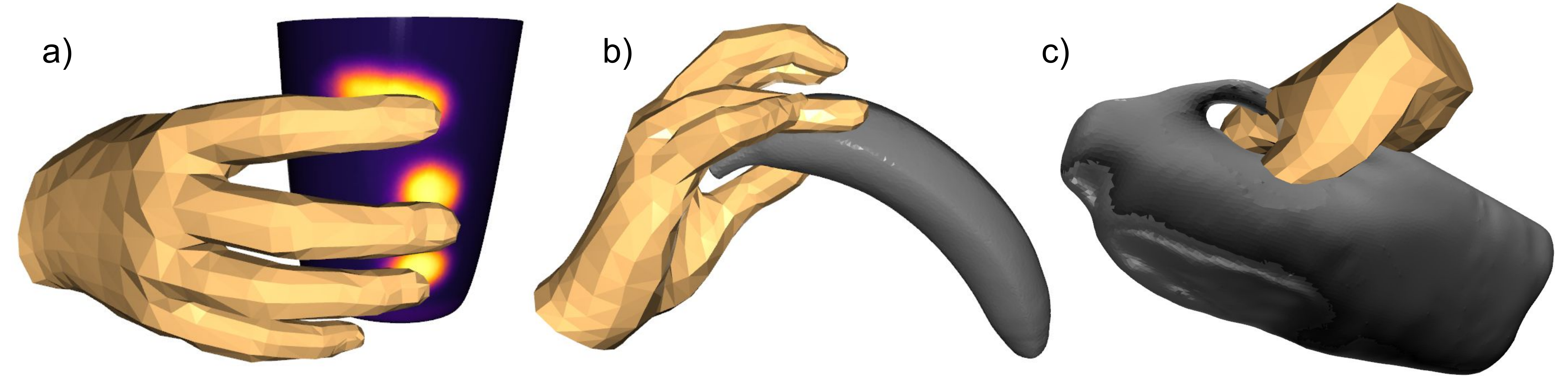}
\end{center}
   \caption{Examples of contact inaccuracy in dataset ground truth annotations: (a) ContactPose \cite{Brahmbhatt_2020_ECCV} (alignment offset), (b) HO-3D \cite{hampali2020honnotate} (hand self-penetration, hand-object gap), and (c) FHAB \cite{garcia2018first} (hand-object penetration).}
   \label{fig:bad_poses}
\end{figure}


\begin{figure*}

\begin{minipage}[b]{0.36\textwidth}

    \centering
    \includegraphics[width=\linewidth]{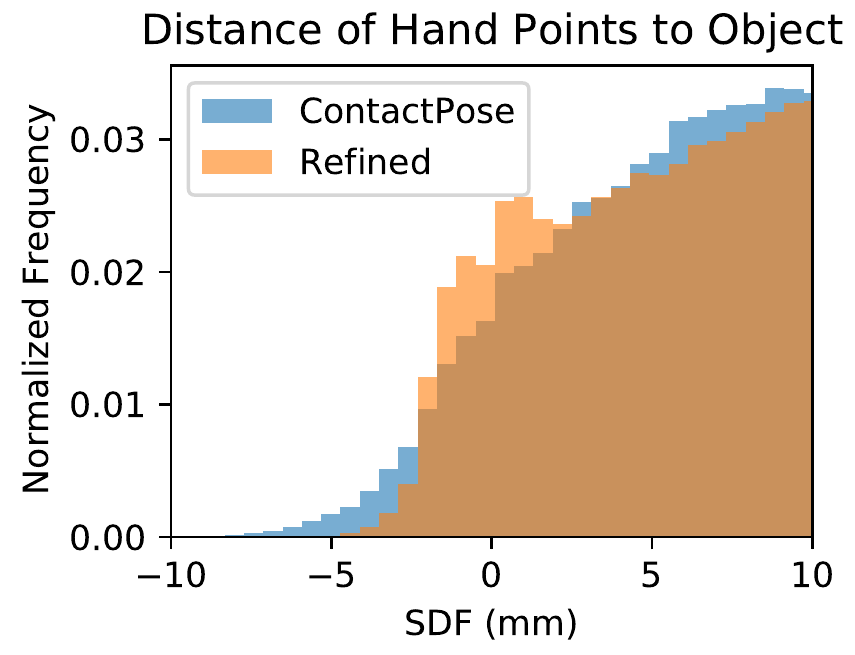}
    \caption{Distance of hand points to object surface, before and after refinement of ContactPose. Note that unrealistic deep interpenetrations (negative) have been mostly eliminated while the fraction of vertices near the surface of the object $[-2, 2]$ mm has increased.
    \label{fig:sdf_fine}}

\end{minipage}
\hfill
\begin{minipage}[b]{0.61\textwidth}

    \centering
    \includegraphics[width=1.0\linewidth]{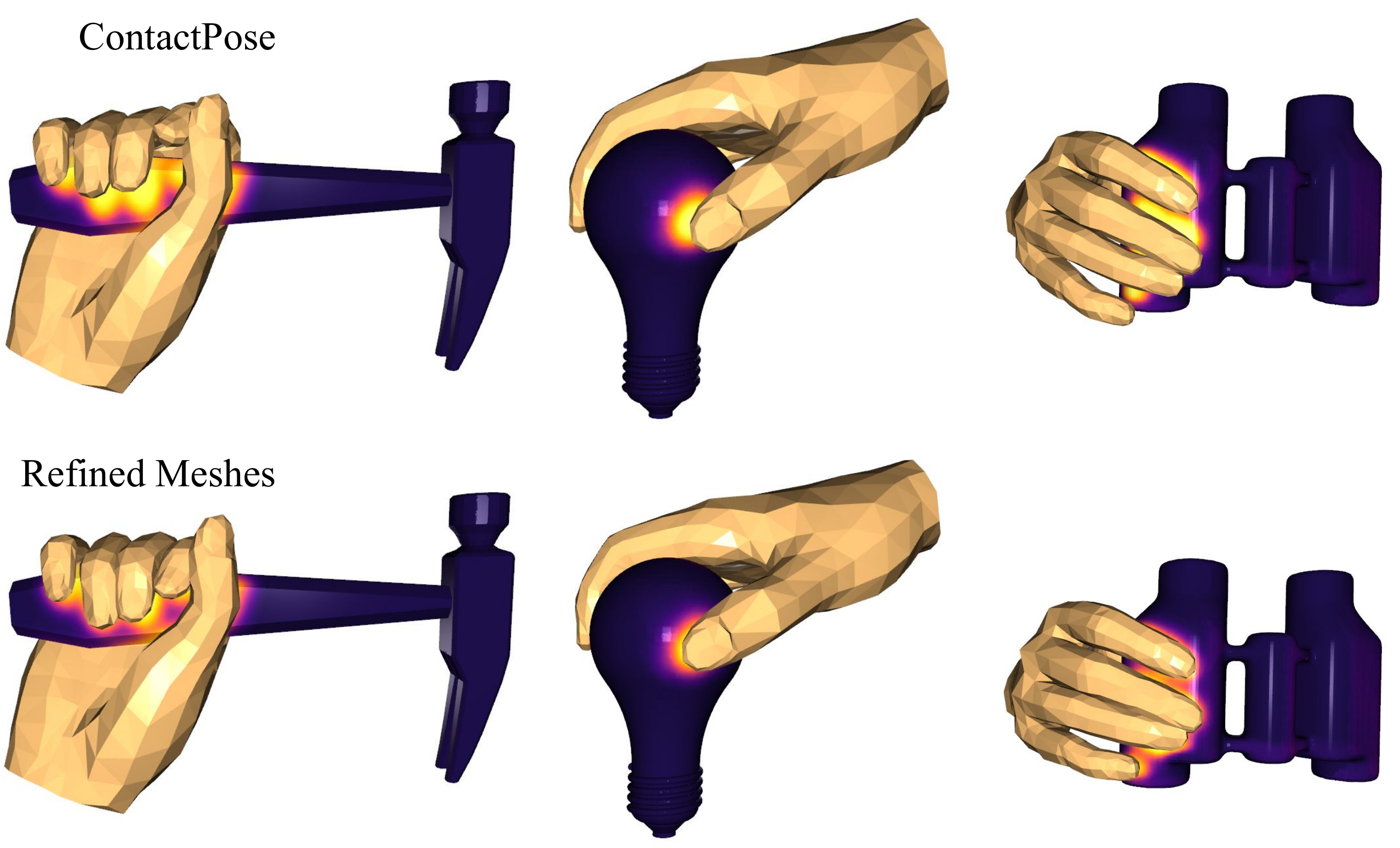}
    \caption{Top: Original meshes from ContactPose with misalignment between hands and contact maps. Bottom: After refinement using \algoName{}.  See \secref{sec:contactpose_results}.
    \label{fig:fine_before_after}}

\end{minipage}

\end{figure*}




\newcommand{\y}{\textbf{\checkmark}}
\newcommand{\n}{$\mathbf{\times}$}
\begin{table*}
\centering
\begin{tabular}{c|c|c|c|c|c|c|c}
    \multirow{2}{*}{\textbf{Dataset}} & \textbf{\algoName} & \textbf{Intersection} & \textbf{MPJPE} & \multicolumn{4}{c}{\textbf{Score (\%) $\uparrow$}}\\\cline{5-8}
    & \textbf{Refinement} & \textbf{Volume} (cm$^3$) $\downarrow$ & (mm) $\downarrow$ & \textbf{Perceptual} & \textbf{Coverage} & \textbf{Precision} & \textbf{Recall}\\\hline
    
    \multirow{2}{*}{ContactPose \cite{Brahmbhatt_2020_ECCV}} & \n & 2.45 $\pm$ 1.99 & - & 30.6 $\pm$ 3.8 & 6.9 & 64.6 & 34.0 \\\cline{2-8}
    & \y & 1.35 $\pm$ 0.90 & 8.06 & \textbf{69.4 $\pm$ 3.8} & 8.9 & \textbf{75.9}  & \textbf{50.0}\\
    
    \hline\hline
    
    Perturbed & \n & 8.46 $\pm$ 16.49 & 79.89 & - & 2.3 & 9.9 & 11.5 \\\cline{2-8}
    ContactPose & \y & 12.83 $\pm$ 8.00 & 25.05 & -  & 19.7 & \textbf{38.7}  & \textbf{54.8} \\
    \hline
\end{tabular}
\caption{Effect of ContactOpt refinement on the ContactPose ground-truth (top 2 rows) and Perturbed ContactPose dataset (bottom 2 rows). The precision and recall scores quantify (\secref{sec:experiments}) agreement with the measured contact map. \algoName improves both perceptual quality and contact agreement.}
\label{tab:contactpose_results}
\end{table*}

\subsection{Refining Small Inaccuracies}\label{sec:contactpose_results}

We use the ContactPose dataset to evaluate the ability of \algoName to improve poses with small inaccuracies. Recent hand-object interaction datasets use a variety of techniques to capture hand and object pose, such as magnetic trackers, multi-view reconstruction from RGB-D cameras, or motion capture systems. Despite using high quality sensors, errors on the centimeter-level are not uncommon (Figure \ref{fig:bad_poses}).

However, when considering the realism of grasps, \textit{millimeters matter}.  Gaps between the hand and object result in unstable grasps and can be visually unsatisfying. Similarly, unrealistic penetration can violate basic assumptions of intact hands and objects. Notably, millimeters of Euclidean error can result in a physically implausible grasp.

\algoName can be used to resolve these types of errors when applied to already high-quality poses provided by dataset annotations. 




\noindent\textbf{Refining ContactPose Dataset Poses}:
Millimeter-scale refinement is demonstrated by refining the ContactPose annotated hand meshes. Rather than estimating target contact using \networkName, the ground truth thermal contact map is used. As ground truth hand contact is not available, hand contact is not used. Table \ref{tab:contactpose_results} and Figure \ref{fig:fine_before_after} show the results of this experiment.

Both contact recall and precision metrics increase, demonstrating that \algoName improves the self-consistency between ground truth contact and mesh poses. Both unwanted contact as well as excess contact are reduced (Figure \ref{fig:sdf_fine}).

However, it is difficult to quantify the holistic quality of a grasp. We perform a perceptual evaluation where human participants choose the most natural-looking grasp. Contact maps are not shown to the participants. As shown in Table \ref{tab:contactpose_results}, participants favored the refined grasps at over a 2:1 ratio. \algoName is able to consistently resolve cases of millimetric penetration or under-shoot and pull the fingers into realistic contact with the object, which is likely noticed by the participants. 

This demonstrates that contact and accurate poses can be used together to achieve higher quality than is possible with pose alone.


\subsection{Refining Large Inaccuracies}

We evaluate the ability of \algoName to improve poses with large inaccuracies in two ways. First, we use perturbed poses from the ContactPose dataset. Second, we use poses estimated from images.

\subsubsection{Refining Perturbed ContactPose} \label{sec:perturbedRefine}

We test the full \algoName pipeline on Perturbed ContactPose (Section \ref{sec:contact_inference}), which contains poses with an MPJPE of $\sim$80 mm. This tests the ability to improve hand poses with large errors. Results are shown in Figure \ref{fig:perturb_before_after} and Table \ref{tab:contactpose_results}.

Despite being initialized from a heavily misaligned hand pose, the pipeline is still able to reduce kinematic error (MPJPE) by almost 70\% and improves perceptual grasp quality. Additionally, the refined meshes are more consistent with the ground truth contact maps, even though they are not provided to the algorithm.

However, some kinematic error remains. Qualitatively, this is because the objects have many valid grasp modes (\ie grasping an apple in any rotation), and it is not possible to recover the correct one from the inaccurate initial pose. Although most refined meshes are visually high quality, often a slight translation results in a large kinematic error.



\subsubsection{Refining Image-Based Pose Estimates}






\begin{figure*}
\begin{center}
   \includegraphics[width=1.0\linewidth]{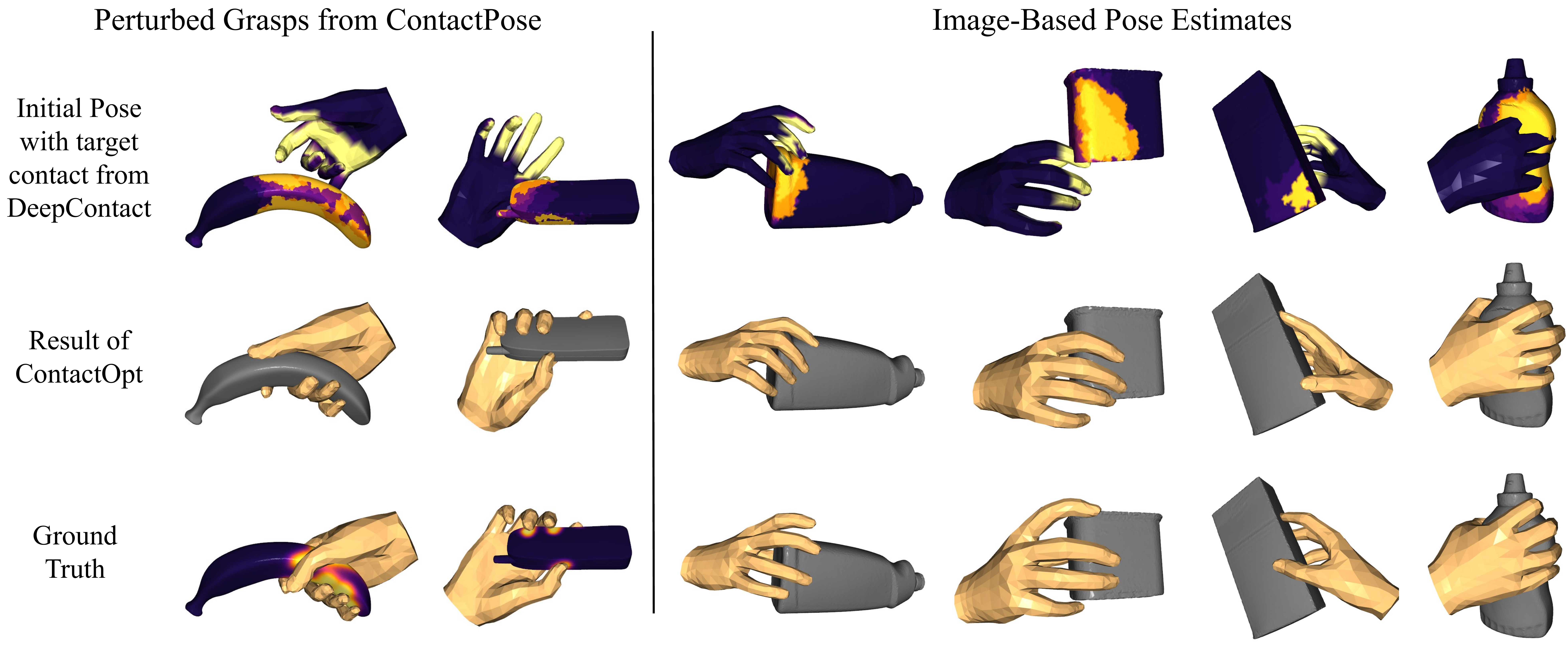}
\end{center}

\caption{Application of ContactOpt to poses from Perturbed ContactPose and image-based pose estimates.
The leftmost column presents an example where the refined grasp is of higher perceptual quality, but as \networkName estimated a different grasp mode, the grasp has high kinematic error. More examples are available in the supplementary material.}
\label{fig:refineHOnnotate}
\label{fig:perturb_before_after}

\end{figure*}

\begin{table*}
\centering
\begin{tabular}{c|c|c|c|c}
    \multirow{2}{*}{\textbf{Method}} & \textbf{Intersection} & \textbf{MPJPE} & \multicolumn{2}{c}{\textbf{Score (\%) $\uparrow$}}\\\cline{4-5}
    & \textbf{Volume} (cm$^3$) $\downarrow$ & (mm) $\downarrow$ & \textbf{Perceptual} & \textbf{Coverage}\\\hline

    Image Pose Estimator~\cite{hasson2020leveraging} & 15.3 $\pm$ 21.1 & 57.7 & \textit{reference} & 4.4 \\\hline\hline
    
    RefineNet (n=3)~\cite{GRAB:2020} & 13.8 $\pm$ 19.0 & 56.3 & 69.6 $\pm$ 3.5 & 5.3 \\\hline 
    RefineNet (n=10)~\cite{GRAB:2020} & 11.6 $\pm$ 18.5 & 64.1 & - & 3.9 \\\hline
    
    ContactOpt (ours) & 6.0 $\pm$ 6.7 & \textbf{48.1} & \textbf{85.2 $\pm$ 2.7}  & 14.7 \\\hline\hline
    
    HO-3D Ground Truth~\cite{hampali2020honnotate} & 1.9 $\pm$ 2.8 & - & - & 2.5 \\\hline
\end{tabular}
\caption{Effect of RefineNet and \algoName algorithms on the hand pose predicted by Hasson et al.~\cite{hasson2020leveraging} on the HO-3D dataset. The perceptual studies compare refined poses against the original image-based estimates. The \algoName refinement achieves the lowest MPJPE and is favored by human evaluators.}
\label{tab:honnotate_results}
\end{table*}

We evaluate \algoName in refining the predictions from an image-based pose estimator. 
In this task, 3D hand and object pose are often estimated using CNNs. For approaches that operate on single-frame RGB images, errors in the multiple-centimeter range are typical, leading to physically implausible grasps.
Note that in this setting, there are no image-based constraints placed on the optimization, thus allowing greater freedom of pose refinement.

We use the baseline pose estimation network from Hasson et al. (2020) \cite{hasson2020leveraging} and retrain it on a training split of the HO-3D dataset. As the network's object predictions are often unstable, the object class and pose are taken from ground truth. Additionally, poses where the ground truth is not in contact are filtered out. More details can be found in the supplementary material.


We demonstrate that \networkName is able to generalize well to new datasets. Despite being trained on the Perturbed ContactPose dataset, it can still improve estimates on HO-3D, which has both different objects and features dynamic grasps. Generally, since hand and object geometry is mostly consistent across datasets, the domain gap is smaller than modalities such as RGB, where learned methods often must be completely retrained. We qualitatively find that \networkName is able to transfer hand contact more reliably than object contact, as the hand representation (MANO) is consistent across datasets.

Results from this task are found in Table \ref{tab:honnotate_results}. Human evaluators favored the refined grasps over the initial grasp estimates by a ratio of almost 6:1. Additionally, the frequency of contact across the hand for the refined grasps (Figure \ref{fig:handContactFreq}) is similar to ground truth frequencies of contact, while the frequency of contact for originally inferred grasps does not resemble normal grasping patterns.

As the dataset contains shapes with many grasp modes (\ie boxes may be grasped anywhere along the edge), \networkName has difficulty predicting the correct grasp location from a low quality inferred grasp. Figure \ref{fig:refineHOnnotate} shows a refined grasp with high perceptual quality but a large MPJPE error metric. Despite this, \algoName is still able to lower the mean kinematic joint error by ~20\%.





\noindent\textbf{Comparing to Baseline Refinement}: We also compare \algoName to a baseline hand pose refinement method. RefineNet \cite{GRAB:2020} is an end-to-end model trained on the GRAB dataset to refine initial coarse grasp proposals. Given a hand and object mesh, the network predicts pose, rotation, and translation updates. As RefineNet is an iterative method, it is benchmarked with 3 iterations (following the paper) and 10 iterations.

\noindent\textbf{Ablating Random Restarts}: The effect of random restarts on kinematic error is shown in Table \ref{tab:restart}. Due to the non-convexity of the optimization objective, performing random initializations with perturbations to translation improves the performance of \algoName.
\begin{table}[h]
\centering
\begin{tabular}{ |c|c|c|c| } 
 \hline
  $n_{restart}$ & 1 & 4 & 8 \\ \hline
  \textbf{MPJPE} (mm) & 53.6 & 51.2 & 48.1 \\ \hline
\end{tabular}
\caption{MPJPE vs number of random restarts, tested on image-based pose estimates. Compare to Table \ref{tab:honnotate_results}}
\label{tab:restart}
\end{table}

%% file: 5-conclusion.tex
\section{Conclusion}

We introduce \algoName, a method to refine coarsely aligned hand and object meshes. \networkName estimates likely contact on both the hand and the object. \meshContactName then estimates contact based on the current mesh pose. The error between these two estimates is used to optimize hand pose to achieve the target contact.

We show that \algoName is able to improve both dataset-quality meshes when ground truth thermal contact is provided, as well as pose estimations from images, even when tested on a novel object set. In our experiments, optimized grasps achieved lower kinematic error and were preferred by human evaluators.





\vspace{18pt}
\noindent\textbf{Acknowledgements}: \small We thank the anonymous reviewers for their comments to improve this work. We also thank Robert Wang, Yuting Ye, Shangchen Han, Beibei Liu, Chengde Wan, Jeff Petkau, and Henry Clever for their advice and discussions.